\documentclass[conference,letterpaper]{IEEEtran}

\IEEEoverridecommandlockouts
\usepackage{subcaption}
\usepackage{booktabs}
\usepackage{amsmath,amsfonts}
\usepackage{algorithmic}
\usepackage{algorithm}
\usepackage{array}
\usepackage{textcomp}
\usepackage{stfloats}
\usepackage{url}
\usepackage{verbatim}
\usepackage{graphicx}
\usepackage{cite}
\usepackage{caption}

\usepackage[hidelinks]{hyperref}
\usepackage{tikz}
\usepackage{amsmath,mathtools}
\usepackage{multirow}

\usepackage{tikz}
\usepackage{tikz-3dplot}
\usepackage{tikzscale}
\tikzset{>=latex}
\tikzset{near start abs/.style={yshift=-3pt, xshift=-6pt}}

\usetikzlibrary{shapes.geometric,calc,fit,positioning} 

\newcommand{\mbf}[1]{\mathbf{#1}}	
\newcommand{\mbs}[1]{\boldsymbol{#1}}
\newcommand{\states}{\mbs{x}}
\newcommand{\dstates}{\dot{\mbs{x}}}
\newcommand{\p}{\mbf{p}}
\newcommand{\R}{\mbf{R}}
\newcommand{\pdot}{\dot{\mbf{p}}}
\newcommand{\omeg}{\mbs{\omega}}

\newcommand\copyrighttext{%
    \small \color{black} Accepted at the 2025 International Conference on Unmanned Aircraft Systems}
\newcommand\copyrightnotice{%
        \begin{tikzpicture}[remember picture,overlay]
            \node[anchor=north,yshift=-0.6cm] at (current page.north) 
            {\color{white}\fbox{\parbox{\dimexpr\textwidth-\fboxsep-\fboxrule\relax}{\copyrighttext}}};
        \end{tikzpicture}%
    }

\begin{document}

\title{RL-based Control of UAS Subject to\\Significant Disturbance\copyrightnotice}

\author{ Kousheek Chakraborty$^{*,1}$, Thijs Hof$^{*,1}$,
Ayham Alharbat$^{1,2}$, Abeje Mersha$^{1}$
\thanks{\textbf{Video}: \href{https://youtu.be/ETapj0mE1kk}{https://youtu.be/ETapj0mE1kk}}%
\thanks{}%
\thanks{* Equal contribution.}
\thanks{This work was supported in part by Horizon Europe CSA project "AeroSTREAM" (Grant Agreement number: 101071270) and Tech for Future project "IFEX Drone" (reference number: TFF 2214)}
\thanks{$^{1}$Smart Mechatronics and Robotics research group, Saxion University of Applied Science, Enschede, The Netherlands.
        }%
\thanks{$^{2}$Robotics and Mechatronics research group, Faculty of Electrical Engineering, Mathematics \& Computer Science, University of Twente, Enschede, The Netherlands.
        }%
\thanks{Corresponding author: Ayham Alharbat \href{mailto:a.alharbat@saxion.nl}{\texttt{a.alharbat@saxion.nl}}}%
}



\maketitle

\begin{abstract}
This paper proposes a Reinforcement Learning (RL)-based control framework for position and attitude control of an Unmanned Aerial System (UAS) subjected to significant disturbance that can be associated with an uncertain trigger signal. The proposed method learns the relationship between the trigger signal and disturbance force, enabling the system to anticipate and counteract the impending disturbances before they occur. We train and evaluate three policies: a baseline policy trained without exposure to the disturbance, a reactive policy trained with the disturbance but without the trigger signal, and a predictive policy that incorporates the trigger signal as an observation and is exposed to the disturbance during training. Our simulation results show that the predictive policy outperforms the other policies by minimizing position deviations through a proactive correction maneuver. This work highlights the potential of integrating predictive cues into RL frameworks to improve UAS performance.

\end{abstract}

\section{Introduction}
Unmanned Aerial Systems (UAS) are increasingly deployed in high-risk environments to perform critical tasks such as infrastructure inspection, search and rescue, and aerial firefighting \cite{drones1010002}. In many of these applications, UAS are equipped with tools or payloads that can induce significant forces and disturbances during operation, potentially destabilizing the system or degrading its performance.

One example of these systems is the IFEX UAS developed by the Smart Mechatronics and Robotics Research Group for assisting firefighters and first responders. The UAS is equipped with a high-powered water cannon, which releases short, targeted, high-velocity water bursts to suppress the fire. However, these bursts generate powerful recoil forces on the floating UAS body, which leads to large deviations from the desired trajectory or, in extreme cases, destabilizes the system. 

\begin{figure}
    \centering
    \includegraphics[width=\linewidth]{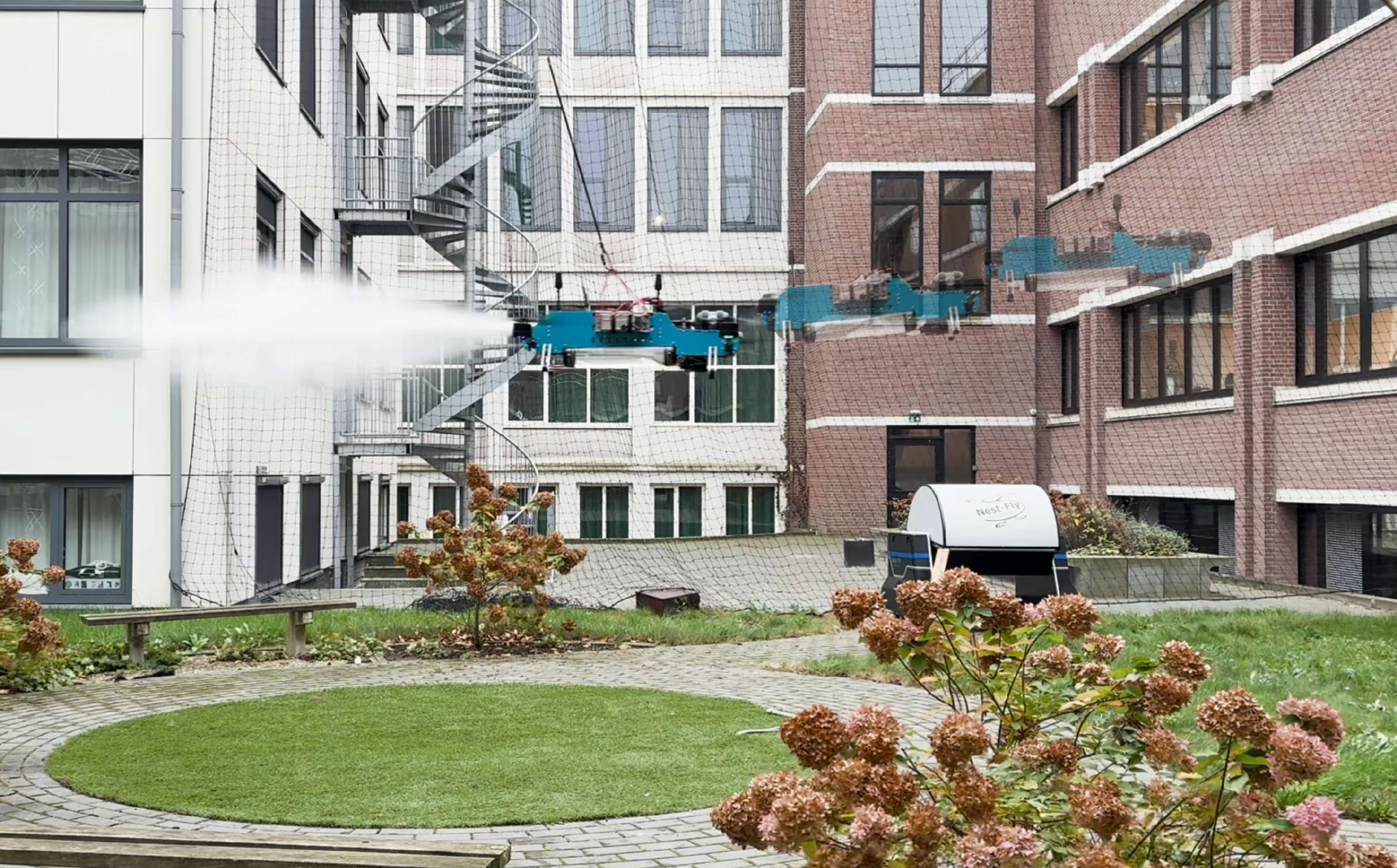}
    \caption{IFEX system firing a high-pressure water burst, illustrating the significant recoil force experienced by the UAS.}
    \label{fig:cover}
\end{figure}

Counteracting this disturbance, caused by the recoil, is a complicated control problem, especially for underactuated systems like quadrotors. 
Traditional reactive trajectory tracking controllers 
try to mitigate these kinds of external disturbances but fall short in addressing the sudden and intense nature of the recoil forces, leading to large tracking errors, as shown in Fig. \ref{fig:cover}.

The operational context of the IFEX UAS is unique because the water cannon activation is manually triggered by the operator, making the disturbance both predictable and repeatable. This predictability allows for proactive stabilization strategies, where the UAS could anticipate the recoil and take corrective actions in advance.

One possible solution is to develop a model of that disturbance and use classical feedforward controllers to counteract the predictable disturbance \cite{tomizuka1987zero}. However, if the disturbance force is applied as a lateral force to the UAS, orthogonal to the rotors' thrust direction, the classical feedforward approach will not be as effective in mitigating this disturbance, due to the under-actuation of the quadrotor \cite{seifried2012two}.

To address these problems, we propose a reinforcement learning (RL)-based control framework that augments the UAS’s observation space to include a trigger signal corresponding to the imminent disturbance. This trigger signal acts as an artificial warning mechanism, allowing the RL algorithm to preemptively counteract the disturbance. Our objective is to investigate the performance of this RL-based approach in addressing the problem of predictable disturbances in aerial robots. Specifically, we aim to test the hypothesis that incorporating a trigger signal into the observation space of an RL-based control framework enables a quadrotor to better anticipate and counteract recoil forces, resulting in improved stability and performance compared to methods that do not use this information.

\textbf{Contributions:} We propose a reinforcement learning-based approach for controlling UAS subject to predictable disturbances. By augmenting the observation space with a trigger signal that informs the agent of an impending disturbance, the system can anticipate and proactively counteract destabilizing forces. We demonstrate the effectiveness of this approach by comparing it with RL policies trained without the trigger signal.

\subsection{Related Work}
Reinforcement learning (RL) has gained significant attention in aerial robotics to solve complex control and planning problems. RL's ability to learn adaptive policies directly from interaction with the environment makes it particularly appealing for tasks involving nonlinear dynamics. One of the first works to demonstrate RL for quadrotor control was presented in \cite{7961277}, where RL was used to track waypoints and recover from challenging initial conditions. Since then, RL has been used with quadrotors to follow high-speed trajectories \cite{doi:10.1126/scirobotics.abg5810} and navigate through cluttered environments while optimizing performance metrics such as speed and path smoothness \cite{10582409}, \cite{Huang2023CollisionAA}. These successes highlight the potential of RL to handle challenging control problems by learning from data rather than relying solely on pre-defined models.

Traditionally, control problems involving disturbance rejection in aerial robotics have been addressed using methods such as PID controllers, geometric controllers \cite{5717652}, and model predictive control (MPC) \cite{Kamel2017}. PID controllers are widely used for their simplicity and reliability in stabilizing drones under small disturbances. Geometric controllers extend this by operating directly on the configuration manifold of the quadrotors \cite{5717652}. MPC optimizes control inputs over a receding time horizon, allowing drones to track trajectories and handle constraints in dynamic environments \cite{Kamel2017}. However, these approaches are reactive in nature, focusing on mitigating the effects of disturbances after they occur. Moreover, their performance is limited when faced with sudden, high-magnitude disturbances, as they often assume disturbances to be random and unpredictable.

Although there is no existing research directly addressing RL for predictable disturbance rejection in aerial robots, the demonstrated success of RL in trajectory tracking and racing tasks suggests its potential to solve similar problems. In these studies, RL policies outperform traditional controllers \cite{doi:10.1126/scirobotics.adg1462} by learning to optimize performance under challenging conditions. However, the absence of a predictive mechanism for handling external disturbances limits their applicability to scenarios where disturbances are predictable.

Our work bridges this gap by using RL to reject predictable disturbances, such as those caused by the release of high-velocity water bursts from a system onboard a UAS. Unlike trajectory tracking or racing, the problem involves preemptive stabilization in response to a predictable impulse disturbance. By incorporating a trigger signal into the RL observation space, we demonstrate that the UAS can anticipate and counteract destabilizing forces. This represents a novel application of RL in aerial robotics, taking advantage of the predictability of the disturbance to achieve stability during operational scenarios.

\section{Modeling}

\subsection{Quadrotor Model}

\begin{figure}[!t]
    \centering
    \includegraphics[width=\linewidth]{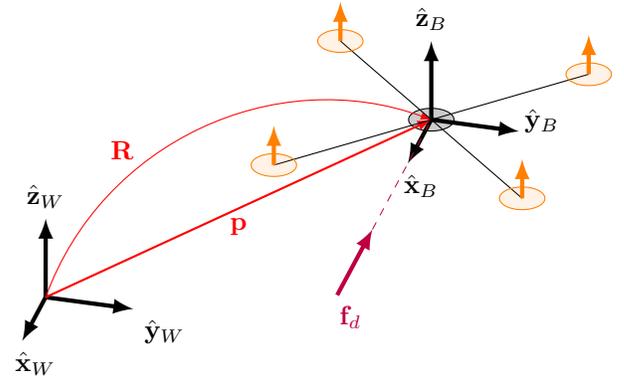}
    \caption{Schematic representation of a quadrotor with its reference frames, and the disturbance force $\mbf{f}_d$ which is applied along the $\hat{\mathbf{x}}_B$ axis.}
    \label{fig-MRAV-framework}
\end{figure}

First, the world inertial frame is defined as $\mathcal{F}_W=O_W\{ \hat{\mathbf{x}}_W,\hat{\mathbf{y}}_W,\hat{\mathbf{z}}_W \}$, while the frame $\mathcal{F}_B=O_B\{ \hat{\mathbf{x}}_B,\hat{\mathbf{y}}_B,\hat{\mathbf{z}}_B \}$ is the body frame, which is fixed to the geometric center of the UAS and aligned with its principal inertia axes, such that the inertia matrix $\mathbf{J} \in \mathbb{R}^{3\times 3}_{>0}$ is a diagonal matrix, as shown in Fig. \ref{fig-MRAV-framework}.

The position of the UAS is denoted as $\mathbf{p} \in \mathbb{R}^{3}$, which is the position of $O_B$ in the world frame. While the attitude of the UAS is expressed using the rotation matrix $\mathbf{R} \in \mathbb{R}^{3 \times 3}$, that is the rotation matrix from $\mathcal{F}_B$ to $\mathcal{F}_W$.

On the other hand, the translational velocity is denoted as $\dot{\mathbf{p}} \in \mathbb{R}^{3}$, expressing the velocity of the body frame with respect to the world frame expressed in the world frame, while the angular velocity $\boldsymbol{\omega} \in \mathbb{R}^{3}$ is the angular velocity of the body frame w.r.t the world frame expressed in the world frame.

The states of the UAS are defined as:
\begin{equation}\label{eq-states-definition}
    \boldsymbol{x} \coloneqq \begin{bmatrix}
    \p^{\top} & \pdot^{\top} & \mathbf{R}^{\top} &  \omeg^{\top} 
    \end{bmatrix}^{\top}
\end{equation}

Then the Newton--Euler equations of motion are:
\begin{equation}
    \dstates = \begin{bmatrix}
        \pdot \\
        - g\, \hat{\mathbf{x}}_W +  \frac{1}{m} \R \left( f_a \hat{\mathbf{z}}_B + \mbf{f}_d  \right) \\
        \R \left[\omeg \right]_{\times} \\
        \mathbf{J}^{-1} \left(  - \boldsymbol{\omega} \times \mathbf{J} \boldsymbol{\omega}  + \boldsymbol{\tau}_a           \right)
    \end{bmatrix}
\end{equation}
where $m \in \mathbb{R}_{>0}$ is the UAS mass, $g$ is the gravitational acceleration, and $f_a \in \mathbb{R}, \boldsymbol{\tau}_a \in \mathbb{R}^{3}$ is the collective thrust and body torques, respectively, generated by the actuators, and $\mbf{f}_d \in \mathbb{R}^{3}$ is disturbance force.

Assuming that the rotors are uni-directional, the actuators' collective thrust and body torques are modeled as:

\begin{equation}
    \begin{bmatrix}
        f_a \\ \mbs{\tau}_a
    \end{bmatrix}
    =
    \begin{bmatrix}
        1 & 1 & 1 & 1 \\
        d_{45} & d_{45} & -d_{45} & -d_{45} \\
        -d_{45} & d_{45} & d_{45} & -d_{45} \\
        c_t & -c_t & c_t & -c_t \\
    \end{bmatrix}
    \left( c_f \, \mbs{\Omega}^{2}  \right)
    \end{equation}
where $\mbs{\Omega} \in \mathbb{R}^{4}$ is the rotational velocities of the rotors, $c_f, c_{t} \in \mathbb{R}^{+}$ is the thrust and drag coefficient of the rotors, and $d \in \mathbb{R}^{+}$ is the distance between the rotor and the geometric center of the quadrotor, such that $d_{45} = d\cos(45^{\circ})$. 
    
\begin{figure}
    \centering
    \includegraphics[width=0.9\linewidth]{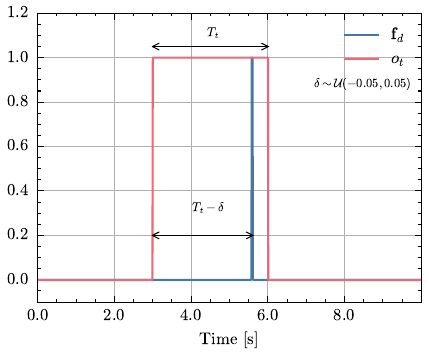}
    \caption{Temporal relation between the impulse disturbance force $\mbf{f}_d$ and the trigger signal $o_t$. Note that the amplitude of the disturbance $\mbf{f}_d$ is normalized for visualization purposes only.}
    \label{fig:impulse-trigger-relation}
\end{figure}
\subsection{Disturbance Modeling}
The impulse disturbance caused by the UAS-mounted water cannon was characterized through a series of experiments on our UAS platform. The water cannon's actuation generates a high-magnitude, short-duration recoil force that sometimes destabilizes the UAS.

The disturbance force, denoted as $\mbf{f}_d$ and expressed in the body frame $\mathcal{F}_B$, was observed to act primarily along the $\hat{\mathbf{x}}_B$ axis of the body frame. The magnitude and duration of the impulse were found to be consistent between repeated shots. This characteristic behavior enabled us to approximate the impulse as a discrete force applied to the UAS's CoM, parameterized by its peak amplitude.

In addition, we incorporated a stochastic noise component to account for slight variations in the recoil dynamics observed during real-world operations. This model was used in the simulations for our reinforcement learning training environments, ensuring realistic and reproducible scenarios that closely mirror experimental observations.

\subsection{Warning Trigger Signal Modeling}\label{sec-trigger-model}
To enable the UAS to counteract the destabilizing effects of the water cannon's recoil, we designed a warning trigger signal, denoted as $o_t \in \mathbb{R}$, to simulate preemptive awareness of the impulse. 

This will eventually be included in the observation space of the policy. Various signal types were explored, such as countdown signals that linearly decrease in amplitude, count-up signals that increase in amplitude, and other temporally correlated signals based on the relative timing of the impulse and the trigger.

Ultimately, we chose a pulse trigger signal because it offered a neutral and unbiased representation of the pre-impulse warning. In this representation, the warning signal is a binary indicator that activates for a fixed duration $T_{t}$. To account for the uncertainty about the relationship between the trigger signal and the disturbance, the disturbance is assumed to occur at a time $T_{t} - \delta$ after the start of the trigger, as depicted in Fig. \ref{fig:impulse-trigger-relation}, where $\delta \sim \mathcal{U}(-0.05, 0.05)$ and $\mathcal{U}(-0.05, 0.05)$ is a uniform distribution between $-0.05$ and $0.05 \, ;s$. 

The value of $T_{t}$ is a hyperparameter for the RL policy since the trigger signal is an artificial signal that can be varied. The effect of $T_{t}$ on the policy will be discussed in Section \ref{sec-results}.

\section{Method}

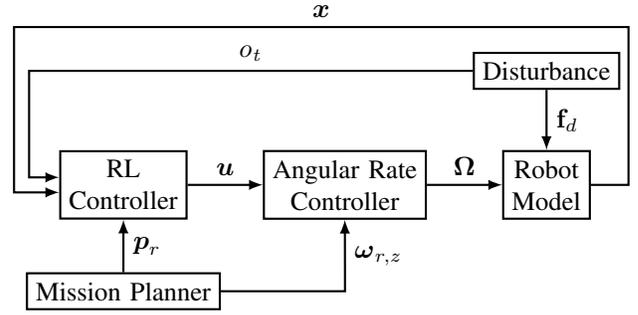
\begin{figure}[!t]
\centering
\begin{tikzpicture}
\node [draw, thick, align=center] (rl) {RL\\Controller};
\node [draw, thick,  below=7mm of rl, align=center] (ref) {Mission Planner};
\node [draw, thick,  right=10mm of rl, align=center] (attc) {Angular Rate\\Controller};
\node [draw, thick,  right=10mm of attc, align=center] (robot) {Robot\\Model};
\node [draw, thick,  above=8mm of robot] (dist) {Disturbance};


\draw[->, thick] (ref.north) -- (rl.south) node[midway, right] {$\mbs{p}_r$};
\draw[->, thick] (ref.east) -| ($(ref.east) + (1.65,0.0)$) -- (attc.south) node[midway, right] {$\omeg_{r,z}$};
\draw[->, thick] (rl.east) -- (attc.west) node[midway, above] {$\mbs{u}$};
\draw[->, thick] (attc.east) -- (robot.west) node[midway, above] {$\mbs{\Omega} $};
\draw[->, thick] (dist.south) -- (robot.north) node[midway, right] {$\mbf{f}_d$};
\draw[->, thick] (dist.west) -| ($(rl.west) + (-0.4,0.1)$) node[pos=0.25, above] {$o_{\mathit{t}}$} -- ($(rl.west) + (0.0,0.1)$);
\draw[->, thick] (robot.east) -| ($(robot.east) + (0.5,2.1)$) -- ($(rl.west) + (-0.6,2.1)$) node[midway, above] {${\states}$} |- ($(rl.west) - (0.0,0.1)$);
\end{tikzpicture}
\caption{Block diagram of the proposed control framework.}
\end{figure}

In this section, we formulate the stabilization of a quadrotor affected by a predictable impulse disturbance as a reinforcement learning (RL) problem. We define this problem as a Markov Decision Process (MDP) $\mathcal{M} = (\mathcal{S}, \mathcal{A}, P, r, \gamma)$ where $ \mathcal{S}$ is the \textit{state space}, $\mathcal{A}$ is the \textit{action space}, $P(s' \mid s, a)$ is the \textit{transition probability}, $r: \mathcal{S} \times \mathcal{A} \to \mathbb{R}$ is the \textit{reward function} and $\gamma \in [0,1]$ is the \textit{discount factor}.
The objective is to learn an optimal policy $\pi^* $ that maximizes the expected cumulative discounted reward:

\begin{equation}
J(\pi) = \mathbb{E} \left[ \sum_{t=0}^{\infty} \gamma^t r_t \right]
\end{equation}
The specific details of the MDP, including the reward function, observation space, and action space, are defined in the following sections.

\subsection{Reward}

We adapted the reward function from \cite{Dionigi_2024} to optimize for smaller position errors and less control effort. The position error, defined as $\mbf{e}_p \in \mathbb{R}^{3}$ is defined as:
\begin{equation}
    \mbf{e}_p = \mbf{p} - \mbf{p}_r
\end{equation}
where $\mbf{p}_r \in \mathbb{R}^{3}$ is the position reference.

The reward for the position error at time-step $k$ is defined as

\begin{equation}
    r_e=(r_x \; r_y \; r_z)^2 \quad \in \mathbb{R}
\end{equation}
where

\begin{equation}
    r_j = max(0, 1-\alpha|\mbf{e}_{p,j}|)
\end{equation}
$r_j$ and $\mbf{e}_{p,j}$ are the $j$-th entries of $r_e$ and $\mbf{e}_p$. The $\alpha$ parameter is introduced to ensure a non-zero reward for $\mbf{e}_{p,j}<\frac{1}{\alpha}$ opposed to just $\mbf{e}_{p,j}<1$. This change was necessary due to the larger position errors caused by the impulse disturbance.

The reward for the control effort is defined as

\begin{equation}
    r_u=-\|\mbf{u}\|^2
\end{equation}
The total reward is calculated as

\begin{equation}
    r=
    \begin{cases}
        r_e + k_u r_u  & ||\mbf{e}_p|| < e_m\\
        -c & \text{otherwise}
    \end{cases}
    \label{eq:reward}
\end{equation}
Where $k_u$ is a weighting parameter to adjust the trade-off of position error and control effort, $c \in \mathbb{R}^{+}$ is a very large positive constant to punish large position errors, and $e_m \in \mathbb{R}$ is the maximum allowed position error distance {such that the episode terminates if it was exceeded}.

\subsection{Observations}
The observations vector $\mbf{o}$ consists of the position error together with the robot's attitude, velocity, angular velocity, and a trigger observation $o_{\mathit{t}}$ for the predictable disturbance, such that:
\begin{equation}
    \mbf{o} \coloneq \begin{bmatrix}
        \mbf{e}_p^{\top} 
        &
        \pdot^{\top} 
        & 
        \mathbf{R}^{\top} 
        &  
        \omeg^{\top} 
        & 
        o_{\mathit{t}}
    \end{bmatrix}
\end{equation}
A history of length $H$ is used to provide the agent with previous observations.

\subsection{Actions}
The agent uses a 3-dimensional action space consisting of a desired collective thrust and two desired angular velocities in $\mathcal{F}_B$ about $\hat{\mathbf{x}}_B$ and $\hat{\mathbf{y}}_B$:

\begin{equation}
    \mbf{u} \coloneq \begin{bmatrix}
        f_a & \omeg_{r,x} & \omeg_{r,y}
    \end{bmatrix}^{\top}
\end{equation}

The agent's actions $\mbf{u}$, together with the desired yaw rate $\omeg_{r,z}$, which is controlled separately, are used as inputs for an angular rate controller 
to obtain the desired individual motor rotor velocities $\mbs{\Omega}$. Note that $f_a$ does not include the gravity compensation component, which is handled by the angular rate controller.


\section{Results} \label{sec-results}

Three policies were trained to evaluate the impact of including the impulse disturbance and the trigger observation during training:

\begin{enumerate}
    \item Nominal: Baseline policy trained without the impulse disturbance or trigger observation.
    \item I-Policy: Policy trained with the impulse disturbance but without trigger observation.
    \item IT-Policy: Policy trained with the impulse disturbance and trigger observation.
\end{enumerate}

No training was performed for the scenario without the impulse disturbance but with trigger observation, as this observation would have no meaningful contribution to the agent.

To evaluate the performance of the agents, an experiment is designed where the quadrotor is hovering at a random position, then after a random time interval, the trigger signal is activated, and the impulse disturbance is applied. The time between the trigger activation and the disturbance is random, as described in section \ref{sec-trigger-model}.

The performance is compared by calculating the position error and the control effort. For the position error, the Root Mean Square Error (RMSE) is calculated at the end of each episode, for each individual axis:

\begin{equation}
    \mathbf{p}_j=\frac{1}{8}\sum_{i=1}^8 \sum_{k=0}^{M} \sqrt{\frac{\mbf{e}_{p,j}(k)^2}{M}}
\end{equation}
Where $M \in \mathbb{N}^{+}$ is the number of time-steps in each episode. These individual RMSE values are averaged over 8 episodes to improve the reliability of the results.

Similarly, the metric for the control effort is defined as:
\begin{equation}
    \Sigma_u=\frac{1}{8}\sum_{i=1}^8 \sum_{k=0}^M\|\mathbf{u}(k)\|
\end{equation}

\subsection{Experimental Setup}
The policies are trained using parallelized simulations in Aerial Gym \cite{kulkarni2023aerialgymisaac}. A custom environment is created to include the impulse disturbance and the corresponding trigger observation. 

The agent is trained using an actor-critic framework. This framework implements two Deep Neural Networks (DNNs), the actor, which learns the optimal control policy, and the critic, which is responsible for evaluating the control policy during training. Both DNNs consist of 3 hidden layers, each with 256 neurons with ELU activations and a final layer outputting a 3-dimensional vector. The input is an $N \cdot H$-dimensional vector, where $N$ is the number of observations for a single time step.

Domain randomization is used to improve robustness and sim-to-real transferability. The randomization is applied to the initial position, orientation, and velocity of the robot. Furthermore, randomization is also applied to the time and force of the impulse disturbance. The reference position is fixed at $\mbf{p}_r=\mbf{0}$.

The physical parameters of the aerial platform that is simulated in training are described in Table \ref{tab:physical_parameters}.

\begin{table}[b]
    \centering
    \caption{UAS Physical Parameters and Disturbance Parameters}
    \begin{tabular}{l c c}
        \multicolumn{3}{c}{\textbf{UAS Properties}} \\
        \midrule
        Mass & $m$ & 3.1 kg \\
        Moment of Inertia & $I_{xx}$ & 0.039 kg$\cdot$m$^2$ \\
        Moment of Inertia & $I_{yy}$ & 0.039 kg$\cdot$m$^2$ \\
        Moment of Inertia & $I_{zz}$ & 0.061 kg$\cdot$m$^2$ \\
        Arm Length & $d$ & 0.28 $m$ \\
        Max Rotor Rotational Velocity & $\bar{\mbs{\Omega}}$ & 800 rad/s \\
        Rotor Thrust Coefficient & $c_f$ & 0.000025 \\
        Rotor Drag Coefficient & $c_{t}$ & 50 \\
        \midrule
        \multicolumn{3}{c}{\textbf{Impulse Disturbance Parameters}} \\
        \midrule
        Impulse Force (x) & $\mbf{f}_{d,x}$ & -950 to -1050 N \\
        Impulse Force (y, z) & $(\mbf{f}_{d,y}, \mbf{f}_{d,z})$ & (0.0, 0.0) N \\
        Trigger Time Before Impulse & $T_{t}$ & 0.5 s \\
        \hline
    \end{tabular}
    \label{tab:physical_parameters}
\end{table}

Training is carried out using the implementation of PPO \cite{schulman2017proximalpolicyoptimizationalgorithms} from RL Games \cite{rl-games2021}. The learning rate $\alpha$ is set to $1e-4$, with lr\_schedule set to "adaptive", allowing the learning rate to change adaptively after every miniepoch depending on the KL divergence at that point. Training is parallelized with 1024 environments running simultaneously. Training is terminated after 1000 epochs. Each episode lasts a maximum of 1000 timesteps, which is equal to 10 seconds in simulation time. Episodes are terminated early when the maximum error is exceeded ($\|\mathbf{e}\|>e_m$). For all three agents, we performed 10 runs with the same training configurations using a fixed set of seeds.

\begin{figure}
    \centering
    \includegraphics[width=0.9\linewidth]{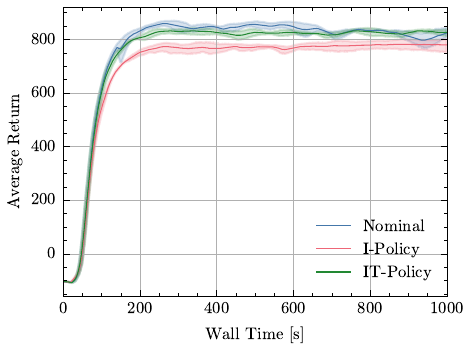}
    \caption{Overview of the average return comparison of the three policies.}
    \label{fig:train_comp}
\end{figure}

As shown in Fig. \ref{fig:train_comp}, all three agents converge, and the IT-Policy outperforms the I-Policy by achieving a higher average return in the same number of training steps. Since the Nominal Policy is not subjected to the disturbance during training, it also performs similarly to the IT-Policy.

\subsection{Timeseries Analysis}

\begin{figure*}
\centering
\begin{minipage}{0.49\textwidth}
  \centering
  \includegraphics[width=\textwidth]{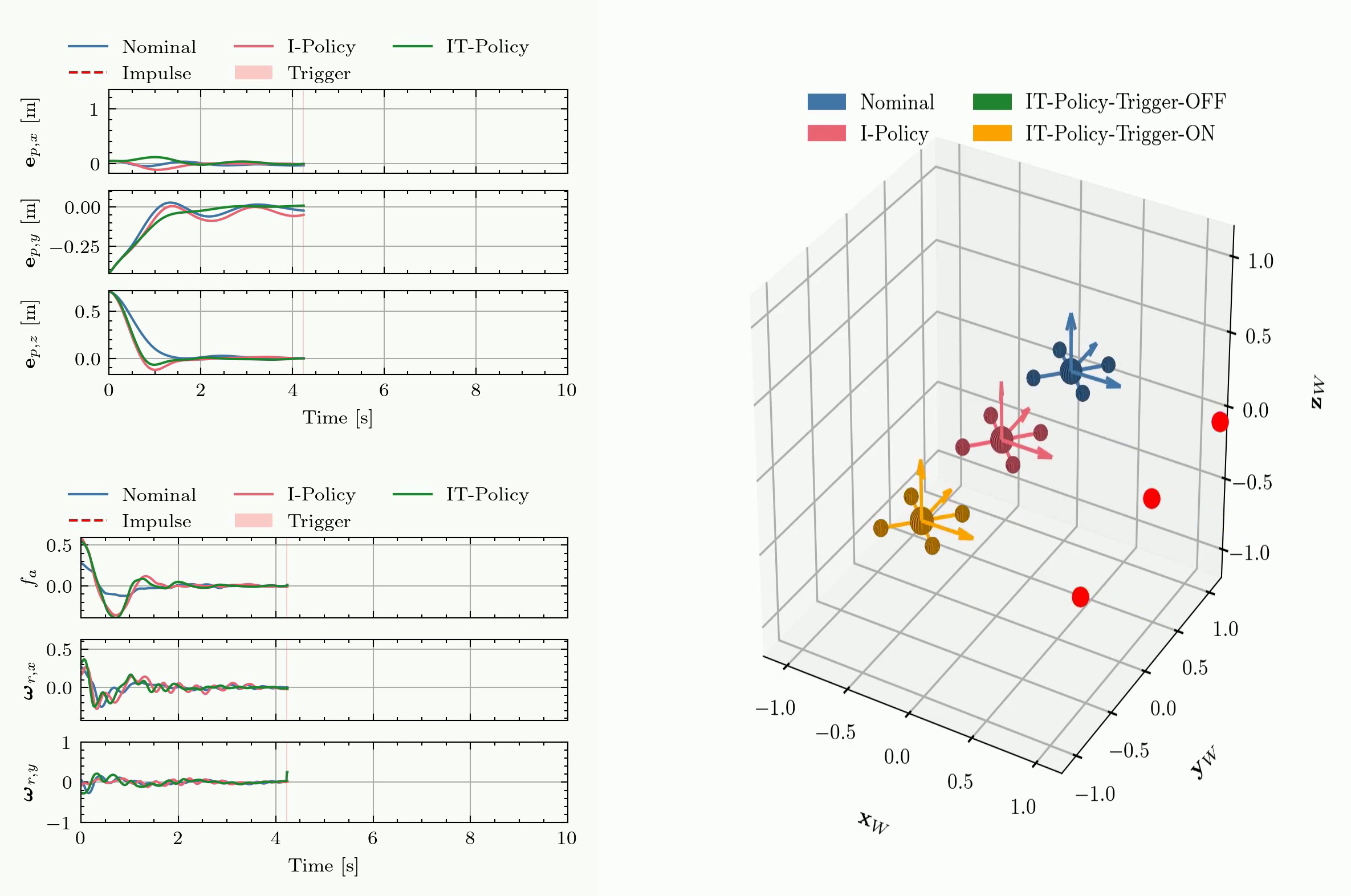}
    \subcaption{}\label{fig:1a}
\end{minipage}%
\hfill
\begin{minipage}{0.49\textwidth}
  \centering
  \includegraphics[width=\textwidth]{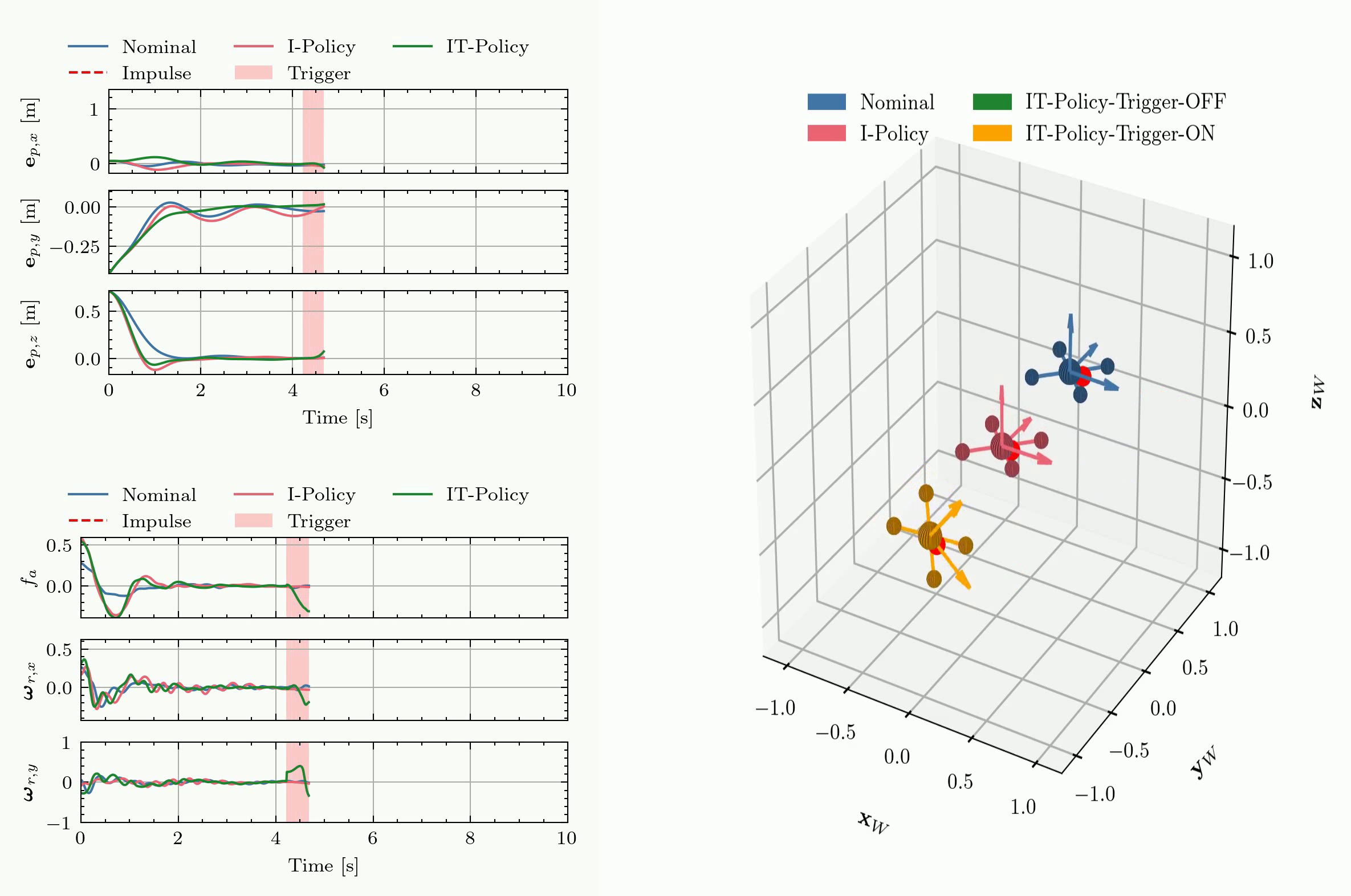}
    \subcaption{}\label{fig:1b}
\end{minipage}%
\\
\begin{minipage}{0.49\textwidth}
  \centering
    \includegraphics[width=\textwidth]{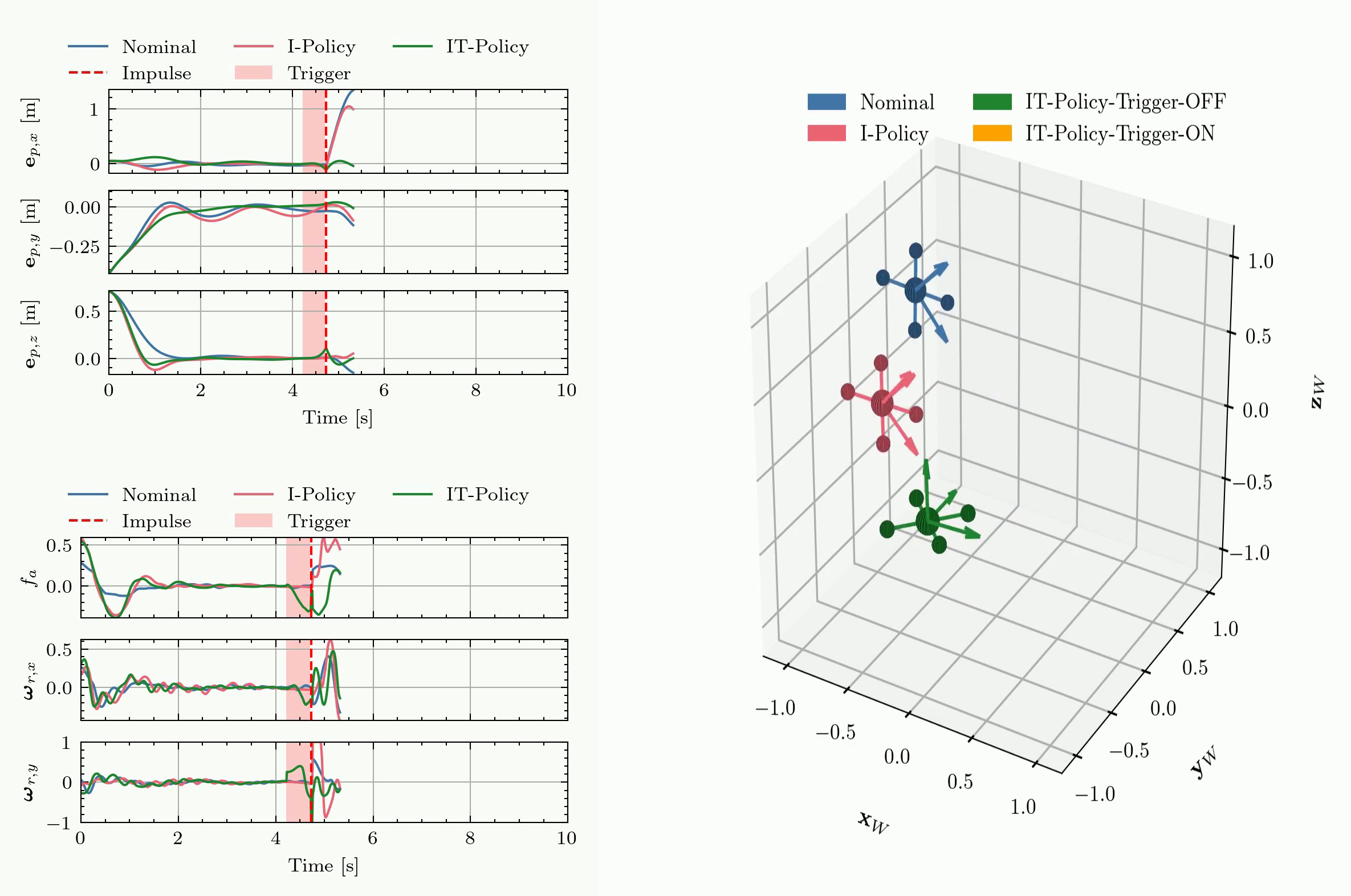}
    \subcaption{}\label{fig:1c}
\end{minipage}
\hfill
\begin{minipage}{0.49\textwidth}
  \centering
  \includegraphics[width=\textwidth]{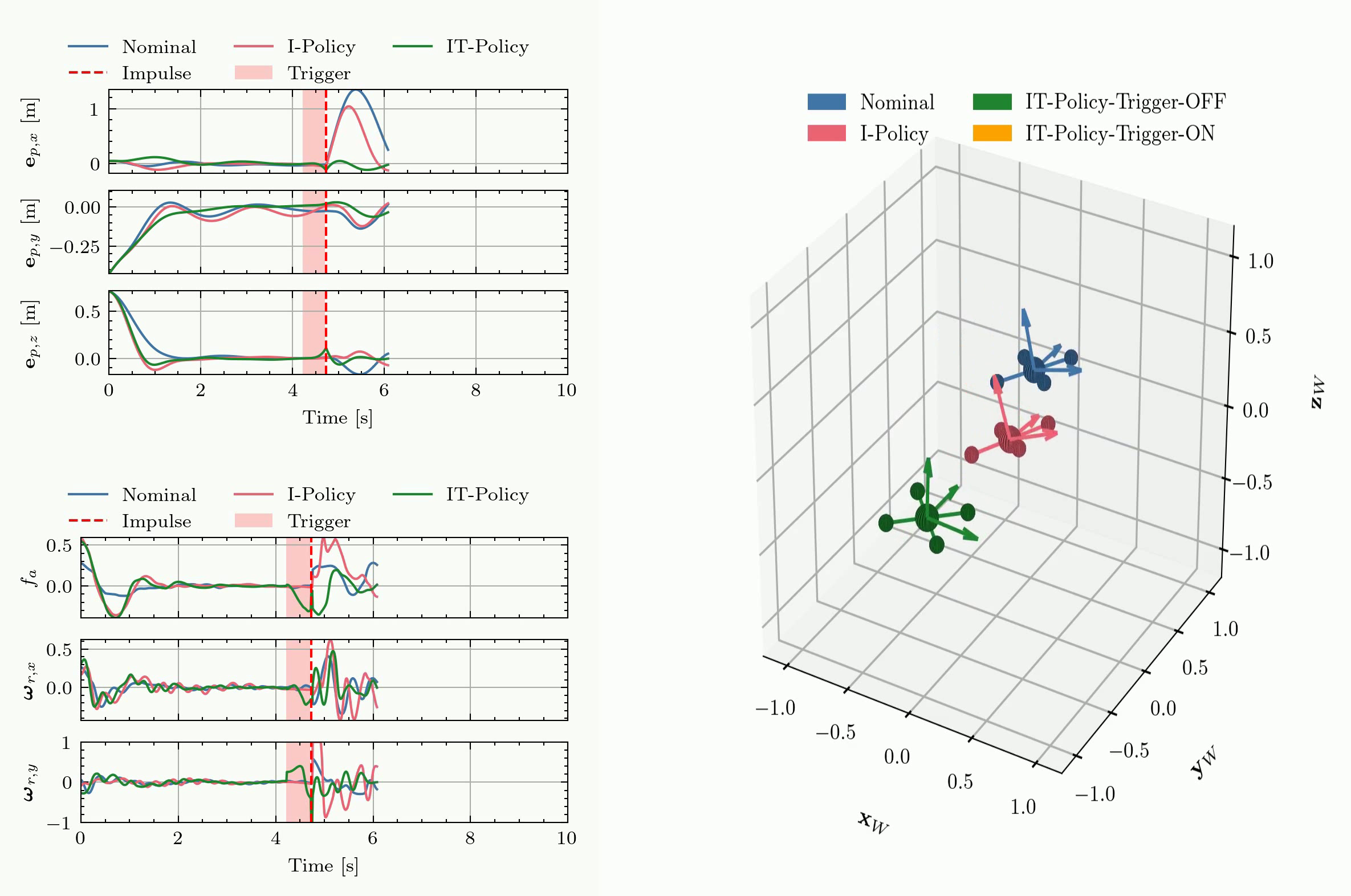}
    \subcaption{}\label{fig:1d}
\end{minipage}
\caption{Snapshots of key moments during a testing episode that combine an animation of three quadrotors controlled by the three policies, and plots of the position errors and control inputs. The impulse disturbance force is visualized using the red dots that approach the quadrotors, and when the dots hit the quadrotors, the disturbance force is applied. Snapshot (a) captures the first timestep after the trigger signal is on. Snapshot (b) captures the timestep right before the impulse disturbance force is applied to the system. Snapshot (c) captures the timestep with the highest deviations in the position error of the three policies. While snapshot (d) shows the oscillations of the nominal and I-Policy before they go back to hovering.} 
\label{fig:1}
\end{figure*}

\begin{figure}
    \centering
    \includegraphics[width=0.9\linewidth]{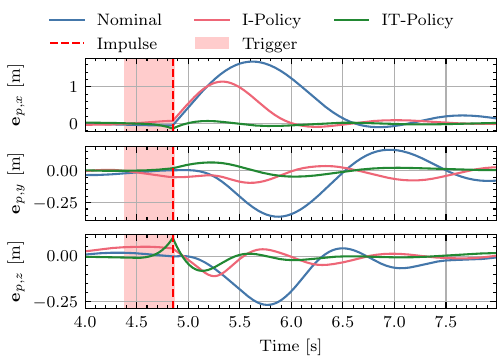}
    \caption{Comparison of the position error along each axis for the three agents, with the following parameters $H=0$, $T_{t}=0.5, \delta=0.02$}
    \label{fig:error_comp}
\end{figure}

The three agents have been evaluated over eight episodes, such that each episode uses the same pseudo-random conditions for all agents. A comparison of the error over time in a single episode for these agents is shown in Fig. \ref{fig:error_comp}, which shows the position error along the x, y, and z axes for the three agents. 

Along the x-axis, where the disturbance occurs, we observe clear differences in how each policy responds to the impulse. The \textbf{Nominal} agent, which was trained without the impulse disturbances and without the trigger signal, deviates significantly from the set point and takes considerable time to recover.

In contrast, the \textbf{I-Policy}, which was trained with the impulse disturbance but without the trigger signal, handles the disturbance more effectively. Although it still deviates from the set point, the policy reacts to minimize the error and return to the target position.

Finally, the \textbf{IT-Policy} demonstrates the best performance. Since it was trained with the impulse disturbance and the trigger signal, it learns to associate the trigger with the imminent disturbance. This anticipation allows the policy to initiate corrective actions early by slightly pitching forward, thus reducing the total deviation from the set point when the disturbance actually takes place, as shown in Fig. \ref{fig:1b}. As a result, the IT-Policy shows the least deviation in comparison to the other agents.

\begin{figure}
    \centering
    \includegraphics[width=0.9\linewidth]{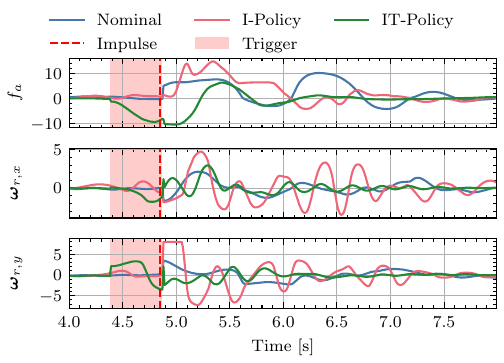}
    \caption{Comparison of the RL actions between the three agents, with the following parameters $H=0$, $T_{t}=0.5, \delta=0.02$}
    \label{fig:control_comp}
\end{figure}

A comparison of the control outputs of the agents is shown in Fig. \ref{fig:control_comp}. The plot displays the desired thrust and angular rates of the three agents over time. The \textbf{IT-Policy} increases $\boldsymbol{\omega}_{r,y}$ at the start of the trigger signal to initiate a preemptive forward tilt, which is combined with a gradual decrease in $f_a$, as shown in Fig. \ref{fig:1b}. The \textbf{I-Policy} shows a large increase in desired $\boldsymbol{\omega}_{r,y}$ at the time of the disturbance, followed by oscillations in both $\boldsymbol{\omega}_{r,x}$ and $\boldsymbol{\omega}_{r,y}$, as shown in Fig. \ref{fig:1c}. The \textbf{Nominal} agent shows a similar increase in $\boldsymbol{\omega}_{r,y}$ and oscillations, but with a smaller magnitude.

\begin{table}[b]
    \centering
    \caption{Comparison between the three agents based on performance metrics: Position tracking errors, and Control Effort.}
    \begin{tabular}{c||ccccc}
                   & $\mathbf{p}_x$  & $\mathbf{p}_y$   & $\mathbf{p}_z$  & $\|\mathbf{p}\|$ & $\Sigma_u$\\
        \midrule
        Nominal  & 0.827 & 0.315 & 0.219 & 0.912 & 148\\
        I-Policy  & 0.278 & \textbf{0.161} & 0.122 & 0.344 & 238\\
        IT-Policy & \textbf{0.106} & 0.173 & \textbf{0.101} & \textbf{0.227} & \textbf{148}\\
    \end{tabular}
    \label{tab:metrics_comp}
\end{table}

\subsection{Impact of History Length and Trigger Duration} 

To investigate the relationship between the trigger signal duration and the length of the observations history, we trained multiple policies with different values of trigger signal duration and history length, and we compared the performance of the policies using the same metrics we defined before, the position error and control effort.

The performance of agents trained using varying values for $H$ and $T_{t}$ are provided in Table \ref{tab:ht_comp}. The agents trained using $H=50$ and $T_{t}\geq1.0$ did not lead to stable flight and, therefore are not included. Overall, the metrics do not indicate an increase in performance with the addition of history to the observations.

The best overall performance is achieved with $H=0$ and $T_{t}=0.5$. Similar performance in terms of position accuracy is achieved by the agents with $H=10$ and $H=50$, but these score significantly lower on the control effort. The lack of improvement by including history may be explained by the fact that the observations already fully define the state of the robot \cite{Dionigi_2024}.

For the best combination of $H$ and $T_{t}$, Table \ref{tab:metrics_comp} shows the metrics defined earlier, collected for the three policies. These metrics show that the IT-Policy is overall the best at minimizing deviations from the set point while requiring the smallest control efforts. Since the disturbance $\mbf{f}_d$ acts primarily along $\hat{\mathbf{x}}_B$ in $\mathcal{F}_B$, the error metric $\mathbf{p}_x$ has the smallest value for the IT-Policy in comparison to the I-Policy and the Nominal Policy.
 
\begin{table}[!t]
    \centering
    \caption{Performance metrics for varying $H$ and $T_{t}$}
    \begin{tabular}{ccccccccc}
        \multirow{3}{*}{$H$} & \multicolumn{8}{c}{Values of $T_{t} [s]$ and Metrics} \\
        & \multicolumn{2}{c}{0.1} & \multicolumn{2}{c}{0.5} & \multicolumn{2}{c}{1.0} & \multicolumn{2}{c}{2.0} \\
        \cmidrule(l){2-9}
        & $\|\mathbf{p}\|$ & $\Sigma_u$ & $\|\mathbf{p}\|$ & $\Sigma_u$ & $\|\mathbf{p}\|$ & $\Sigma_u$ & $\|\mathbf{p}\|$ & $\Sigma_u$ \\
        \midrule
        0   & 0.31 & 229 & \textbf{0.23} & \textbf{148} & 0.33 & 212  & 0.33  & 230\\
        10  & 0.28 & 304 & 0.23 & 430 & 0.24 & 388  & 0.33  & 362\\
        50  & 0.28 & 373 & 0.23 & 346 & $\times$ & $\times$  & $\times$ & $\times$\\
    \end{tabular}
    \label{tab:ht_comp}
\end{table}

\section{Conclusions}
In this paper, we presented an RL-based approach for mitigating predictable impulse disturbances in UAS. By incorporating a trigger signal into the observation space, our proposed framework enables the UAS to anticipate and counteract these disturbances to improve stability compared to reactive control strategies. Simulation results from our experiments demonstrated the effectiveness of this approach.

Looking ahead, several directions for future research can be explored. First, real-world validation of the proposed system will be required to determine the Sim2Real transferability of our framework to physical systems. Additionally, extending this approach to other types of aerial robots and dynamic systems with different predictable force profiles could broaden its applicability across various domains, such as autonomous manipulation, aerial logistics, and infrastructure maintenance, leading to a more reliable and generalizable solution for disturbance rejection in aerial systems.

\bibliographystyle{IEEEtran}
\bibliography{main.bib}

\end{document}